\documentclass[sigconf]{acmart}

\copyrightyear{2023}
\acmYear{2023}
\setcopyright{rightsretained}
\acmConference[MADiMa '23]{Proceedings of the 8th International Workshop on Multimedia Assisted Dietary Management}{October 29, 2023}{Ottawa, ON, Canada}
\acmBooktitle{Proceedings of the 8th International Workshop on Multimedia Assisted Dietary Management (MADiMa '23), October 29, 2023, Ottawa, ON, Canada}
\acmDOI{10.1145/3607828.3617798}
\acmISBN{979-8-4007-0284-6/23/10}
\AtBeginDocument{%
  \providecommand\BibTeX{{%
    \normalfont B\kern-0.5em{\scshape i\kern-0.25em b}\kern-0.8em\TeX}}}

\usepackage{makecell}
\usepackage{array}
\usepackage{balance}
\newcolumntype{P}[1]{>{\centering\arraybackslash}p{#1}}
\begin{document}

\title{Muti-Stage Hierarchical Food Classification}

\author{Xinyue Pan}
\orcid{0000-0002-3791-9315}
\email{pan161@purdue.edu}
\affiliation{%
  \institution{Purdue University}
  \city{West Lafayette}
  \state{Indiana}
  \country{USA}
  \postcode{47906}
}

\author{Jiangpeng He}
\orcid{}
\email{he416@purdue.edu}
\affiliation{%
  \institution{Purdue University}
  \city{West Lafayette}
  \state{Indiana}
  \country{USA}
  \postcode{47906}
}

\author{Fengqing Zhu}
\orcid{}
\email{zhu0@purdue.edu}
\affiliation{%
  \institution{Purdue University}
  \city{West Lafayette}
  \state{Indiana}
  \country{USA}
  \postcode{47906}
}



\begin{abstract}
    Food image classification serves as a fundamental and critical step in image-based dietary assessment, facilitating nutrient intake analysis from captured food images. However, existing works in food classification predominantly focuses on predicting 'food types', which do not contain direct nutritional composition information. This limitation arises from the inherent discrepancies in nutrition databases, which are tasked with associating each 'food item' with its respective information. 
    Therefore, in this work we aim to classify food items to align with nutrition database. To this end, we first introduce VFN-nutrient dataset by annotating each food image in VFN with a food item that includes nutritional composition information. Such annotation of food items, being more discriminative than food types, creates a hierarchical structure within the dataset. However, since the food item annotations are solely based on nutritional composition information, they do not always show visual relations with each other, which poses significant challenges when applying deep learning-based techniques for classification. To address this issue, we then propose a multi-stage hierarchical framework for food item classification by iteratively clustering and merging food items during the training process, which allows the deep model to extract image features that are discriminative across labels. Our method is evaluated on VFN-nutrient dataset and achieve promising results compared with existing work in terms of both food type and food item classification.
\end{abstract}





\maketitle

\section{Introduction}
Image-based dietary assessment, which involves analyzing nutrient and energy intake from food images captured by an individual, is becoming increasingly prevalent~\cite{shao2021_ibdasystem}. With the ubiquity of mobile devices, many people find it useful to snap pictures of their meals to track dietary intake and monitor adherence to a healthy eating regimen \cite{SAMAD20222,Coughlin2015}. In addition, image-based dietary assessment is crucial for healthcare applications\cite{allegra2020}.

\begin{figure}[t]
    \centering
    \includegraphics[width=0.99\linewidth]{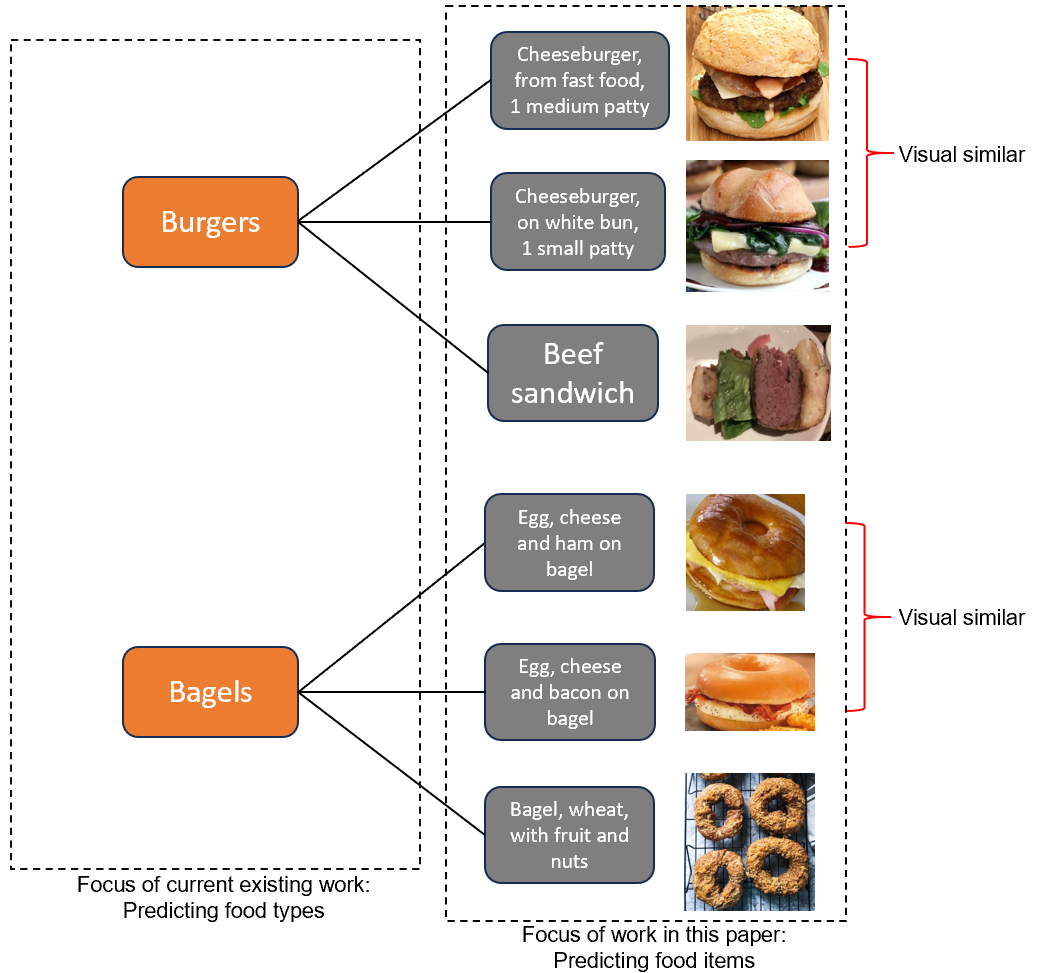}
    \caption{The hierarchical structure in food categorization is depicted in two parts. The left part illustrates the task of food classification in existing work, which focuses on predicting food types. The right side shows the focus of our work: predicting specific food items, each of which is uniquely associated with a particular nutritional composition. A major challenge in classifying food items is that they do not always exhibit visual relations; that is, visually similar images could belong to different food items.}
    \label{fig:hier}
\end{figure}

A vital component of image-based dietary assessment is food image classification, which aims to predict the food consumed in an eating occasion image\cite{boushey2017, he2021end}. In this paper, we establish that a food can be annotated using both its food type and specific food items, as illustrated in Figure \ref{fig:hier}. Food types, which are typical classes (\textit{i.e.,} Apple, Bagels, Burgers, etc.) found in datasets such as Food-101 \cite{bossard14} and UEC-256 \cite{Kawano2014AutomaticEO}, do not carry any associated nutritional composition information. Food items (\textit{i.e.,} Cheeseburger from fast food, Cheeseburger on a white bun, Beef sandwich, etc.) do possess linked nutritional composition information. Because the annotations for food items are more discriminative compared to those for food types, as shown in Figure \ref{fig:hier}, we consider food items as subcategories of food types, thereby forming a hierarchical structure within the dataset. While extensive research has been conducted to enhance the accuracy of food image classification based on food types \cite{Mao2021ImprovingDA, mao2020, he2021, he2020}, such approaches cannot make classifications based on food items. Therefore, the current food classification results are not directly applicable for dietary assessment, even when the volume of food is estimated~\cite{shao2021_toward, shao2023endtoend}. This is because the primary objective of dietary assessment is to perform a nutritional analysis, and the existing classification data lacks corresponding nutritional composition information such as the energy per 100 grams of food. The inability to predict food items stems from the absence of datasets annotated with food items. The lack of such datasets, which would pair each food image with a food item, can be attributed to the domain-specific expertise required for such annotation.

In this paper, we bridge this gap by linking the VFN dataset \cite{mao2020} to nutritional composition information from the USDA Food and Nutrient Database for Dietary Studies (FNDDS) using a USDA food code identifier, as in \cite{lin2023}. In this way, each food image is matched with a food item in FNDDS, which corresponds to a specific USDA food code and is in turn linked to its respective nutritional composition information. Hence, the dataset, named \textbf{VFN-nutrient}, presents a two-level hierarchical structure with food types at the top level and food items at the bottom level. Ultimately, our goal is to conduct food image classification based on food items in the dataset.

While most recent works employ Convolutional Neural Networks (CNNs) \cite{Albawi2017,karen14,He_2016,zagoruyko16,szegedy15} for hierarchical-based image classification \cite{an2021,he2022hierarchical,KOLISNIK2021115195,guo2018}, they assume that labels are visually related to each other at each hierarchical level. However, the food item labels in the VFN-nutrient dataset, which are built based on nutritional composition, do not always exhibit visual correlations. Different food items could be visually similar, as shown in Figure \ref{fig:hier}. This makes it challenging to apply CNN-based models to learn image features directly using food items as labels, because a CNN model learns features visually, and in our case, there is a lack of visual correlation across food items. Additionally, there is a class imbalance issue among food items, causing classification results to be biased towards food items that contain more images. To address these issues, we propose creating visual relations among labels by merging food items iteratively during the training phase through a clustering method and updating image features extracted by the CNN model. We introduce multi-stage hierarchical transfer learning to train and update the model and mitigate the class imbalance issue.



The main contributions of our work can be summarized as follows:
\vspace{-0.7mm}
\begin{itemize}
    \item Unlike most existing works that focus on predicting food types, we aim to predict food items, which include nutritional composition information.
    \item We introduce the VFN-nutrient dataset, which contains nutritional composition information for each food image. 
    \item We propose an end-to-end food item classification system that iteratively merges visually similar food items and employ multi-stage hierarchical transfer learning for improved classification.
\end{itemize}

\begin{figure*}[t]
    \centering
    \includegraphics[width=0.99\linewidth]{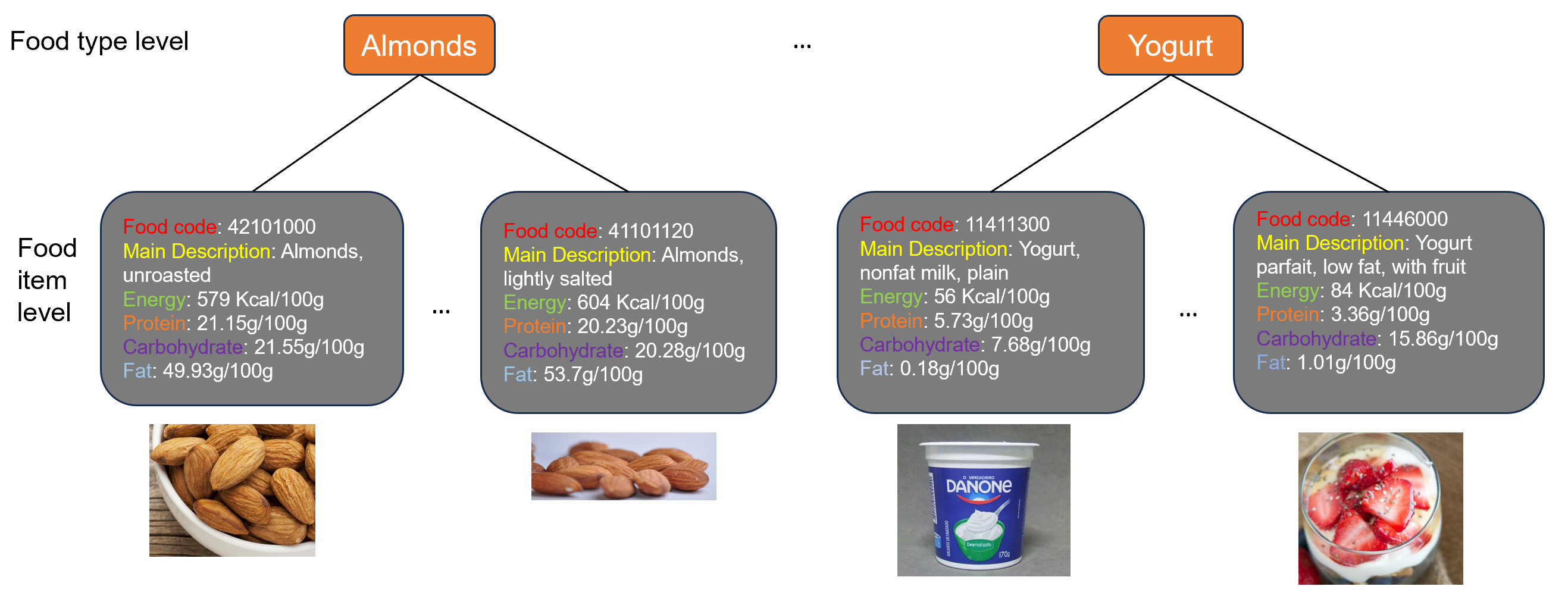}
    \caption{Structure of VFN-nutrient dataset}
    \label{fig:dataset}
\end{figure*}

\section{Related Work}
\subsection{Food image classification}
Many contributions have been made in the field of food image classification from various perspectives.

\noindent\textbf{Improving Food Image Classification Performance on Food Types:} The major challenge of food image classification is the higher inter-class similarity and intra-class diversity compared to other general objects~\cite{mao2020, peng2023selfsupervised, fu2023conditional}. S. Abdulkadir \textit{et al.} enhanced classification performance by concatenating deep features from different models such as VGG~\cite{karen14}, ResNet~\cite{He_2016}, Wide ResNet~\cite{zagoruyko16}, and InceptionV3~\cite{szegedy15} for food type classification. They observed a noticeable improvement over previous classification performance~\cite{Abdulkadir2019}. D. T. Nguyen \textit{et al.} combined local appearance and structural features, specifically integrating non-redundant local binary pattern features and encoding the spatial relationships between interest points, to improve food image classification performance~\cite{NGUYEN2014242}. R. Mao \textit{et al.} proposed a visual hierarchy structure and employed multitask learning to enhance classification accuracy on food types in the VFN dataset~\cite{mao2020}. This approach was further refined by integrating nutritional information into the hierarchy~\cite{Mao2021ImprovingDA}. However, these works still focus on food type classification while the nutritional composition information used is generalized for each food type and does not correspond to each individual food image.

\noindent\textbf{Food Image Classification Under Special Problem Settings} Existing work also attempt to approach the food classification issue through the lens of continual lifelong learning~\cite{ILIO, he2021, he2023longtailed, raghavan2023online, he2022_expfree}, while the work in~\cite{Horiguchi2018PersonalizedCF, pan2022_madima} aims to address the issue from the perspective of personalized food image classification, focusing on food images that appear sequentially over time. Additionally, few-shot learning~\cite{liu2021,Arslan2022,Javier2022}, fine-grained classification~\cite{Jiang2020, Wu2022, mao2020} and long-tailed classification~\cite{he2022long, gao2022dynamic_LTingredient, he2023singlestage} have also been extensively explored in this area.

\noindent\textbf{Datasets Proposed for Food Image Classification:} Several works have focused on proposing food image datasets, such as Food-101~\cite{bossard14}, Food-2K~\cite{min2023}, ISIA-500~\cite{min2020}, and UEC-256~\cite{Kawano2014AutomaticEO}. These datasets are collected through various methods and often focus on specific regions. The aim of introducing these datasets is to enhance the generalization ability of trained models. 

However, all of these existing works fall short in linking classification results to nutritional composition information, a crucial requirement for achieving the objectives of image-based dietary assessment.

\subsection{Hierarchical image classification}
In various contexts, images are associated with different labels, ranging from coarse to fine categories. As such, hierarchical classification is particularly suitable for these scenarios, and numerous important contributions have been made in this field.

\noindent\textbf{Methods Proposed in Hierarchical Image Classification:} H. Long \textit{et al.} proposed a hierarchical feature fusion method that classifies target labels by training each level of the hierarchy separately and then fusing the features from each level for final predictions~\cite{he2022hierarchical}. G. An \textit{et al.} introduced a hierarchical transfer learning method that initially learns the top level of the hierarchy and then transfers the model to the bottom level to make predictions on those categories\cite{an2021}. Additionally, \cite{KOLISNIK2021115195} developed a conditional CNN network for multi-label image classification, which is also related to hierarchical classification. Furthermore, \cite{guo2018} presented a CNN-RNN model to classify leaf categories in a hierarchical structure, leveraging the CNN's capability for feature extraction and the RNN's strength in optimizing the classification of both coarse and fine labels. Y. Zhou \textit{et al.} employed multi-task learning to simultaneously learn labels at different levels of the hierarchy\cite{ZHOU2022108449}.

\noindent\textbf{Datasets Used in Hierarchical Image Classification:} Publicly available hierarchical image datasets like CIFAR-10~\cite{Alex} and ImageNet~\cite{Deng2009} provide training data for hierarchical image classification. In ImageNet~\cite{Deng2009}, the presence of multiple objects in an image contributes to the hierarchical structure, as each image can be assigned multiple labels. For CIFAR-10~\cite{Alex}, certain specific categories are grouped into super-categories, thereby forming a hierarchical structure within the dataset.

However, existing works in hierarchical image classification often assume that classes at the bottom level have visual relationships with each other. Therefore, these methods are not directly applicable to our scenario, where there is limited visual correlation between classes.

\section{Dataset}

\begin{figure*}[t]
    \centering
    \includegraphics[width=0.99\linewidth]{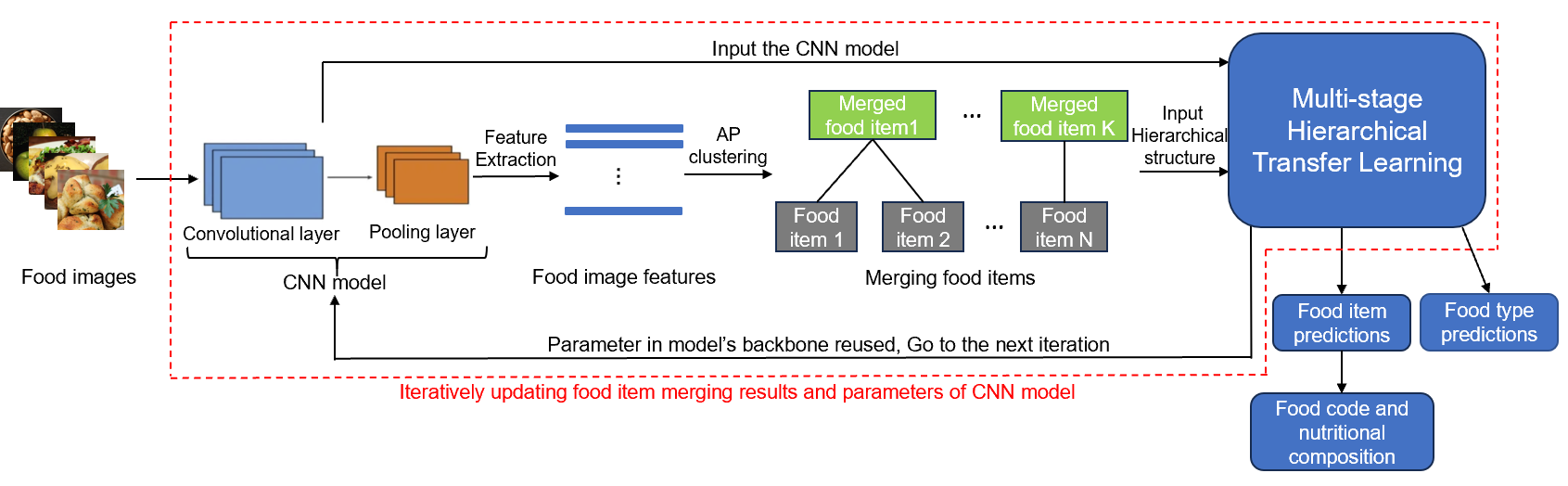}
    \caption{Overview of our proposed method: $N$ represents the total number of food items available for classification. We first extract image features using a CNN model. Second, we merge food items iteratively via Affinity Propagation (AP) clustering, resulting in $K$ merged food items. Third, we train the CNN model using a multi-stage hierarchical transfer learning approach. The trained CNN model is then fed into the next iteration for another round of food item merging, allowing us to update the merging results iteratively. Finally, the iterative process is halted if either the number of iterations reaches 5 or there is no further decrease in the validation loss for predicting food items. At this point, we make inferences on the food items and obtain their nutritional composition information.
}
    \label{fig:method}
\end{figure*}

In this paper, we introduce the VFN-nutrient dataset, which is an extension of the original VFN dataset~\cite{mao2020}. This new dataset comprises 74 food types selected based on the 'What We Eat In America' (WWEIA) survey. The dataset construction follows the methodology presented in~\cite{lin2023}. To annotate food items in each image within the VFN dataset, four experts from the nutrition science team were divided into two independent groups, and they rigorously reviewed all the food images to assign specific food codes from the FNDDS 2017–2018 database to each one. The purpose is to cross-verify each other's work, thereby minimizing subjective errors and improving the reliability of the annotation. Any discrepancies in annotation from two groups were resolved through additional rounds of review involving additional two experts. The aim of the review process was to correct errors, improve categorization, and establish a reliable link to the FNDDS for nutritional analysis.

Each food code corresponds to nutritional composition information and can be considered a food item in the dataset. Consequently, the dataset encompasses 15K images and 945 food items, which belong to 74 different food types. Each image has two labels: one indicating the food type and another specifying the food item.





The structure of the VFN-nutrient dataset is illustrated in Figure \ref{fig:dataset}. The dataset employs a 2-level hierarchical structure: 'food types' make up the top level, while 'food items' constitute the bottom level. The 'food types' are based on the original food classes introduced in the VFN dataset\cite{mao2020}.
Each food item in the VFN-nutrient dataset is associated with a specific food code from the FNDDS, as well as its corresponding nutritional composition information based on a 100g food sample. Because the nutritional composition data allow the food items to be more discriminative relative to each other, we consider food items as subcategories of food types. 


In the dataset, each food item is associated with nutritional composition information based on a 100g food sample. This implies that if we can accurately estimate the weight of the food shown in the image, we can directly calculate the nutrient value for each composition. Given that the ultimate aim of image-based dietary assessment is nutrient analysis based on captured food images, this dataset significantly advances the field toward achieving its intended goal.


However, annotating food items based solely on nutritional composition presents a unique set of challenges for food image classification. Specifically, food items do not necessarily exhibit visual correlations with one another, complicating the task of feature learning for CNN models. In this scenario, the features learned by a CNN model may not be discriminative enough to differentiate between various labels. In this paper, we address this challenge as it specifically relates to food image classification within the dataset. For the experiments to be conducted in subsequent sections, the VFN-nutrient dataset is divided into training, validation, and testing subsets using a 7:1:2 ratio.


\section{Method}
\begin{figure}[t]
    \centering
    \includegraphics[width=0.99\linewidth]{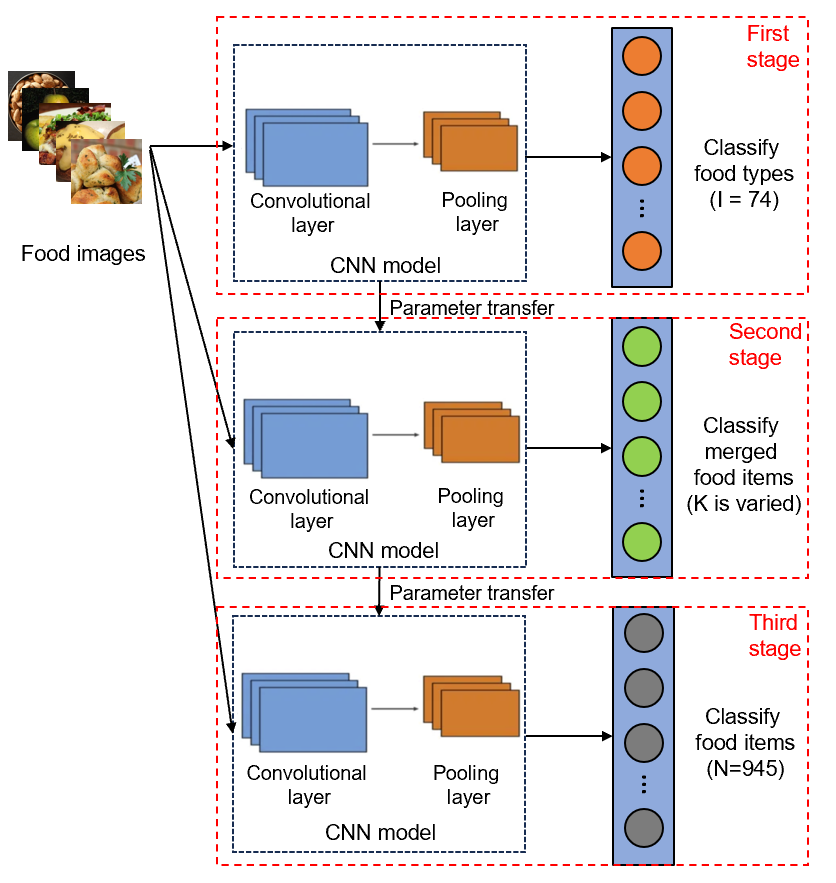}
    \caption{In the multi-stage hierarchical transfer learning approach, the initial CNN model is trained from scratch using food types as labels and is linked to a layer of dimension $I = 74$. The parameters from this model's backbone are then reused in the second stage to build a new CNN model that is connected to a layer of dimension $K$ (variable) and trained with merged food item labels. Subsequently, these parameters are utilized once again in the third stage to create another CNN model that is connected to a layer of dimension $N = 945$ and trained using individual food item labels. The backbone parameters from this final model are then carried forward to the next iteration, continuing another cycle of hierarchical transfer learning. \textit{Note: Labels for merged food items can change with each iteration.}
}
    \label{fig:method_transfer}
\end{figure}

In this section, we introduce a framework designed to classify food items, as illustrated in Figure \ref{fig:method}. The framework aims to tackle two primary issues: the lack of clear visual correlations among different food items and the issue of class imbalance in the dataset. To address these challenges, we introduce an additional level of hierarchy between 'food types' and 'food items'. This is accomplished by iteratively clustering visually similar food items and updating the Convolutional Neural Network (CNN) model through multi-stage transfer learning during the training phase. During the validation and testing phases, the trained model in multi-stage transfer learning stage is applied directly for food image classification.


The method is composed of two main components during the training phase: (1) Employing clustering method to merge food items, (2) Utilizing multi-stage hierarchical transfer learning method to update CNN model and make classification on food items.

\subsection{Merging Visual Similar Food Items}
\label{sec:merge}
To merge food items visually, we first use a CNN model to extract image features. To capture the distribution of image features within each food item, we concatenate the mean and variance feature vectors and feed them as input into a clustering method to merge food items. This input can be represented as \( f = [m_n, v_n] \), where \( m_n \) denotes the mean feature vector for food item \( n \), and \( v_n \) represents the variance feature vector for food item \( n \).

While our approach can accommodate various clustering methods, we specifically utilize Affinity Propagation (AP) \cite{frey2007} in this paper because it does not require a predefined number of clusters. The method operates by transmitting 'messages' among food items; these messages reflect the suitability of one food item to serve as a representative for another. This process updates iteratively across other pairs of food items until convergence. The input to Affinity Propagation consists of the mean and variance features of each food item, concatenated and denoted as $f$. $f_x$ and $f_y$ represent the input features for food items $x$ and $y$, respectively. $s(x, y)$ represents the similarity between food items $x$ and $y$, calculated based on the negative Euclidean distance between their input features. The clustering method utilizes the following equations:

\begin{equation}
    s(x,y) = -||f_x - f_y||
\end{equation}

\begin{equation}
    r(x,y) = s(x,y) - max_{y'=y}[a(x,y') + s(x,y')]
\end{equation}
\begin{equation}
    a(x,y) = min[0, r(y,y) + \sum_{x'\in \{x,y\}}r(x',y)]
\end{equation}

where $r(x,y)$ indicates the suitability of food item $y$ to be the exemplar for food item $x$, and $a(x,y)$ represents the accumulated evidence that food item $x$ should select food item $y$ as its exemplar. The method eventually outputs the food items that are clustered together. We then merge those that are in the same cluster under the same food type to maintain the hierarchical structure. The merged food items are denoted as $[m_1, m_2,..., m_K]$, where $K$ denotes the number of merged food items and $N > K$, given $N$ food items.

The merging process is iterative and continues as the image features undergo refinement due to updates in the CNN model. This iteration persists until one of two conditions is met: either the number of iterations reaches five or the validation loss (i.e., loss on the validation set within the dataset) for food item classification ceases to decrease. This methodology confers several key benefits to our approach. First, the image features learned across different labels become more discriminative owing to the visual correlations among the merged food items. Second, the end-to-end system enables the CNN model to adapt its learning from different sets of merged food items at different iterations, due to the variability in merging outcomes in each iteration. This means that distinct image features can be learned and refined during each iteration. Lastly, since many food items are represented by a limited number of training images, the merging process serves to mitigate the issue of class imbalance during the training phase. 

\subsection{Multi-stage Hierarchical Transfer Learning}

To exploit the hierarchical structure created and to address the class imbalance issue in food items, we adopt multi-stage hierarchical transfer learning. This approach allows us to transfer the knowledge acquired from the top level (food types) to the bottom level (food items). Figure \ref{fig:method_transfer} outlines the pipeline of our multi-stage hierarchical transfer learning approach. Our methodology encompasses a three-stage, iterative training process.

\begin{itemize}
    \item First stage: In the initial iteration, we employ a CNN model that is pretrained on ImageNet\cite{Deng2009}. This model is then connected to a fully connected layer with a dimension of \(I = 74\), utilizing food types as labels. In subsequent iterations, we retain the parameters in the backbone of the CNN model trained in the previous iteration. A new CNN model is constructed and connected to a fully connected layer with a dimension of \(I = 74\) from scratch, and it is trained using food types as labels.
    
    \item Second stage: The parameters from the backbone of the CNN model trained in the first stage are reused to construct a new CNN model and make it connected to a fully connected layer with a dimension of \(K\), where \(K\) varies in different iterations based on the updated food merging results. This model is then trained using merged food items as labels.
    
    \item Third stage: Leveraging the parameters from the backbone of the CNN model trained in the second stage, we construct a new CNN model. This model is connected to a fully connected layer with a dimension of \(N = 945\) and is trained using food items as labels.
\end{itemize}

At each training phase, we utilize the cross-entropy loss as our objective function, formulated as follows:
\begin{equation}
    L = - \sum_{i=1}^N y_{l,i} \log(p_{l,i})
\end{equation}
Here, $l$ denotes the stage at which the model is being trained. $y_{l,i}$ represents the ground truth label $i$ at stage $l$, and $p_{l,i}$ indicates the confidence score for predicting label $i$ at stage $l$.

The CNN model, refined through this multi-stage hierarchical transfer learning process, is then employed in the food item merging process to generate new merging results, as described in section \ref{sec:merge}. Subsequently, the parameters from the model's backbone are transferred to initiate the next iteration of the multi-stage hierarchical transfer learning process. This cycle continues until either the validation loss on food item classification ceases to decrease in subsequent iterations, or until a maximum of 5 iterations is reached to mitigate the risk of overfitting. The CNN model trained in the final iteration of the third stage is ultimately used to make predictions on food items and to retrieve the corresponding nutritional composition information. For experimental comparison with related works, the saved model from the first stage of the last iteration is used to predict food types.


\section{Experiment}
We evaluate our proposed method based on average classification accuracy of predicting food items and mean absolute error in terms of nutritional composition on predicted food items and compare it with other related works on VFN-nutrient dataset. 

\subsection{Experimental setup}
To train the CNN model, we partition the VFN-nutrient dataset into training, validation, and testing sets using a 7:1:2 ratio. The training set is utilized for model training, while the validation set is employed for determining whether the currently trained model should be saved for subsequent testing. The testing set is reserved for final model evaluation. In both our proposed method and related works, the ResNet-50 model\cite{He_2016} serves as the backbone for the CNN architecture. For optimization, we use the Adam optimizer\cite{kingma2017adam} with an initial learning rate of 
0.0001, complemented by a cosine annealing scheduler\cite{loshchilov2017sgdr}. Each stage within the multi-stage hierarchical transfer learning process is trained for 15 epochs. After each iteration, we decrease the initial learning rate by a factor of 
0.8. The training process terminates either after 5 iterations or if the validation loss fails to decrease in a subsequent iteration.

\textbf{Methods for comparison:} 
Our method integrates an end-to-end food merging system to establish visual correlations and employs a multi-stage hierarchical transfer learning approach to continually update the CNN model. This facilitates learning across food types, merged food items, and individual food items. We evaluate our proposed framework by comparing with baseline and existing hierarchical based image classification work including:

\begin{itemize}
\item \textbf{Flat-CNN}\cite{He_2016}: Utilizes a CNN model pretrained on the ImageNet dataset\cite{Deng2009} to train image features with target classes as labels. Inference is then performed on these target classes.

\item \textbf{Visual Hierarchy with Multitask Learning (VHML)}\cite{mao2020}: Establishes a visual hierarchical structure by clustering target classes based on image features extracted using a pretrained Flat-CNN model. It then employs multitask learning to simultaneously classify both clustered and target classes.

\item \textbf{Integrated Hierarchy with Multitask Learning (IHML)}\cite{Mao2021ImprovingDA}: Forms an integrated hierarchy by clustering target classes based on both nutritional composition information and visual features. It then applies multitask learning to concurrently classify both the clustered and individual target classes.

\item \textbf{Hierarchical Feature Fusion (HFF)}\cite{he2022hierarchical}: Trains the CNN model based on labels from each hierarchical level separately. It then concatenates the image features extracted from each trained model for the final classification of target classes.

\item \textbf{Hierarchical Transfer Learning (HTL)}\cite{an2021}: Trains the CNN model first with the top-level classes in the hierarchy as labels and then retrains it using the bottom-level classes as labels for the final classification.
\end{itemize}

We adapt these methods to fit our specific use-case, enabling them to predict individual food items rather than just food types.

\begin{table}[t]
\begin{center}
\caption{Average classification accuracy on predicting food items for different methods: The related work primarily is focusing on predicting food types. We re-implement these methods to fit into our secnario and make them predict food items} \label{tab:result_classification}
\begin{tabular}{c|c}
\hline
Methods & Average classification accuracy(\%)\\
\hline
\textbf{Flat-CNN}\cite{He_2016} & 46.35\\
\textbf{VHML}\cite{mao2020} & 47.42\\
\textbf{IHML}\cite{Mao2021ImprovingDA} & 47.59\\
\textbf{HFF}\cite{he2022hierarchical} & 48.70 \\
\textbf{HTL}\cite{an2021} & 49.55 \\
\hline
\textbf{Ours} & \textbf{50.67}\\
\hline
\end{tabular}
\end{center}
\end{table}

\begin{table}[t]
\begin{center}
\caption{Average classification accuracy on predicting food types for different methods} \label{tab:result_classification2}
\begin{tabular}{c|c}
\hline
Methods & Average classification accuracy(\%)\\
\hline
\textbf{Flat-CNN}\cite{He_2016} & 74.66\\
\textbf{VHML}\cite{mao2020} & 74.95\\
\textbf{IHML}\cite{Mao2021ImprovingDA} & 75.13\\
\textbf{HFF}\cite{he2022hierarchical} & 75.28\\
\hline
\textbf{Ours} & \textbf{75.83}\\
\hline
\end{tabular}
\end{center}
\end{table}

\begin{figure*}[t]
    \centering
    \includegraphics[width=0.99\linewidth]{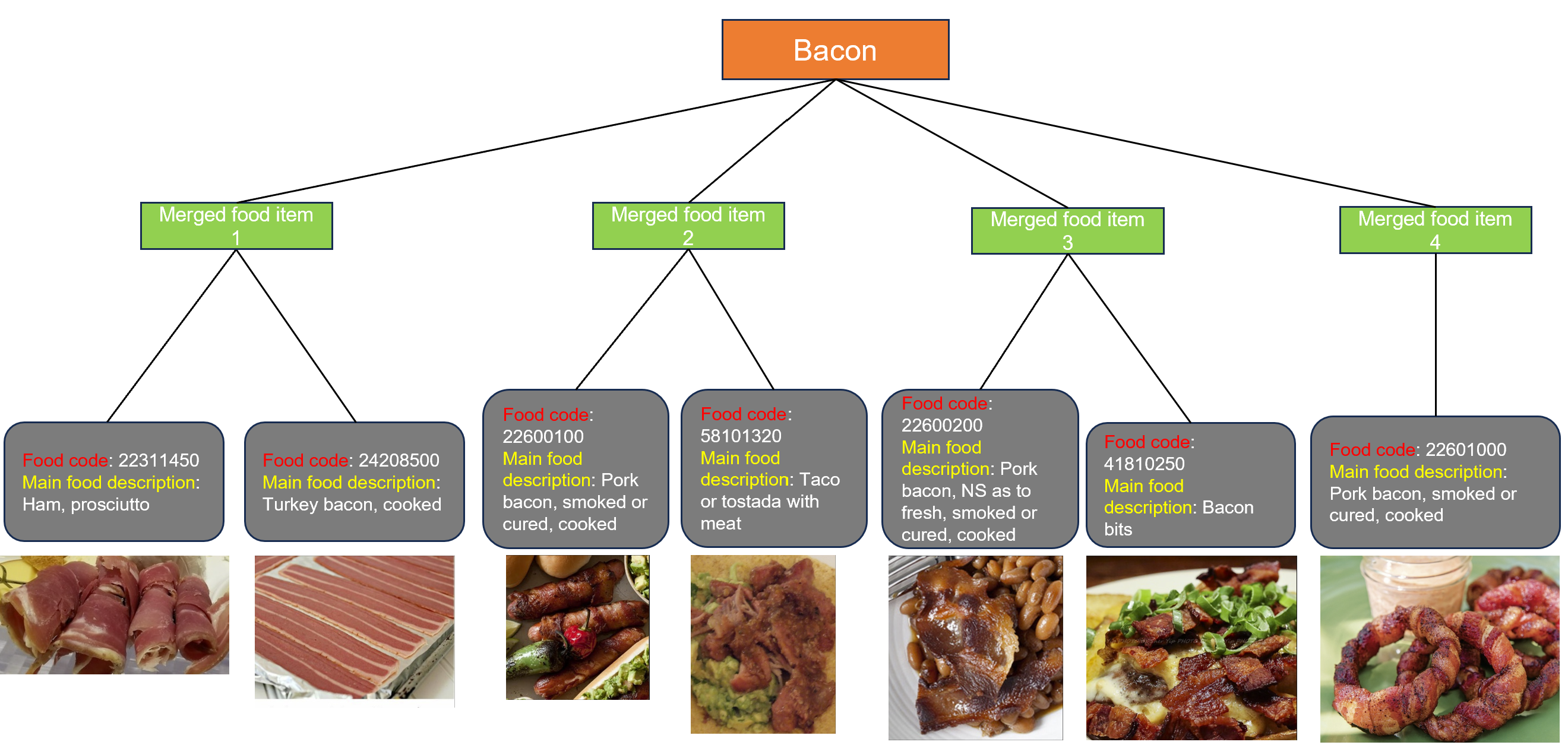}
    \caption{Example of food item merging results}
    \label{fig:qual_item}
\end{figure*}

\subsection{Experiment results}
\subsubsection{Average classification performance on predicting food items}
Table \ref{tab:result_classification} shows the average classification accuracy for predicting food items across different methods. Our method outperforms all other techniques due to several innovative strategies. Specifically, our method creates visual relations by merging food items, which allows the model to learn more distinct image features across labels compared to learning directly with food items as labels. Importantly, this merging process is updated iteratively as the image features are refined, enabling the model to learn different features in each iteration. This aspect distinguishes our work from \textbf{VHML}\cite{mao2020} and \textbf{IHML}\cite{Mao2021ImprovingDA}, where the clustering results are static and do not adapt as the model improves.

In addition to this, our method employs an iterative, multi-stage transfer learning approach. After each iteration where the model finishes learning from food items as label, it reverts to learning from food types as label in the next iteration. Any performance gains achieved during this phase enhance the model's subsequent learning when it switches back to food items as labels again. This dynamic updating sets our method apart from \textbf{HTL}\cite{an2021}, which does not retrain the model at the top level of the hierarchy.Nevertheless, achieving improvements in food item classification remains challenging due to severe inter-class similarity, intra-class dissimilarity, and a low number of training images among various food items. Our proposed method partially mitigates this issue, but it remains a tough problem to fully resolve.

Moreover, multi-stage hierarchical transfer learning in our method not only improves classification accuracy on food items but also on food types, as demonstrated in Table \ref{tab:result_classification2}. Unlike \textbf{HTL}\cite{an2021}, which transfers the backbone parameters from the top-level hierarchy to the bottom level without further iteration (making its performance on food type classification equal to \textbf{Flat-CNN}\cite{Albawi2017}), our method continues to iterate, enhancing its performance. As a result, we do not include \textbf{HTL} in our comparison for food type classification.

\subsubsection{Nutrition analysis}
We utilize the Mean Absolute Error (MAE) as an evaluation metric to assess performance concerning the nutritional composition of food items. Introduced in \cite{Mao2021ImprovingDA}, this metric is particularly useful for understanding the error in terms of nutritional composition on predicted food items. Specifically, a prediction is considered a 'better mistake' if the nutritional composition of the predicted food item closely resembles that of the target food item. The formula for calculating MAE is given by:

\begin{equation}
    MAE = \frac{1}{N} \sum_{i=0}^{N-1} |A_i - \hat{A_i}|
\end{equation}

In this equation, $A_i$ represents the nutritional value per 100g of the predicted food item, and $\hat{A_i}$ represents the nutritional value per 100g of the target food item. $N$ denotes the total number of testing images. We report the MAE in terms of macro nutritional components---energy, carbohydrates, fats, and proteins---as shown in Table \ref{tab:result_mae}.

\begin{table}[t]
\begin{center}
\caption{Mean absolute error in terms of nutritional composition per 100g food sample} \label{tab:result_mae}
\begin{tabular}{P{1.8cm}|P{1cm}|P{1cm}|P{1.8cm}|P{1cm}} \hline
Methods & \makecell{Energy\\(Kcal)} & \makecell{Protein\\(g)} & \makecell{Carbohydrate\\(g)} & \makecell{Fat\\(g)}\\
\hline
\textbf{Flat-CNN}\cite{Albawi2017} & 39.36 & 2.39 & 5.31 & 3.20\\
\textbf{VHML}\cite{mao2020} & 37.69 & 2.18 & 4.90 & 3.08\\
\textbf{IHML}\cite{Mao2021ImprovingDA} & 37.74 & 2.18 & 4.88 & 3.05\\
\textbf{HFF}\cite{he2022hierarchical} & 34.89 & 2.02 & 4.47 & 2.87 \\
\textbf{HTL}\cite{an2021} & 34.98 & 2.06 & 4.48 & 2.85 \\
\hline
\textbf{Ours} & \textbf{33.24} & \textbf{1.98} & \textbf{4.31} & \textbf{2.72}\\
\hline
\end{tabular}
\end{center}
\end{table}

Our method also demonstrates superior performance in terms of nutritional composition errors per 100g food sample compared to existing works. Specifically, our approach is designed to make "better mistakes," meaning that even if the model misclassifies a food item, the predicted item's nutritional profile closely resembles that of the target item. This is a significant advantage, as it ensures a higher degree of accuracy in the nutritional information provided, despite potential misclassifications. 

\subsubsection{Clustering performance at different iterations of proposed method}
To substantiate the efficacy of our iterative method for updating both the merged food items and the backbone of the CNN model, as delineated in Section~\ref{sec:merge}, we examine the clustering results obtained after each iteration using two established metrics: the Silhouette Score~\cite{Shahapure2020} and the Davies-Bouldin Index~\cite{Singh2020}.

The Silhouette Score~\cite{Shahapure2020} serves as an indicator of the distinctness of the clusters relative to each other. Its value ranges from -1 to 1, with higher values representing more distinct and thus better clusters. Conversely, the Davies-Bouldin Index~\cite{Singh2020}, ratio of intra-cluster to inter-cluster distance, offers an inverse interpretation. Lower values of this index imply better clustering performance, as they indicate clusters that are both tight and well-separated.

\begin{table}[t]
\begin{center}
\caption{Clustering performance at in the first 5 iterations of our proposed method} \label{tab:result_clustering}
\begin{tabular}{c|c|c}
\hline
\makecell{Iteration \\Number} & \makecell{Silhouette \\score} & \makecell{Davies \\Bouldin \\index}\\
\hline
1 & 0.036 & 2.059\\
2 & 0.177 & 1.685\\
3 & 0.216 & 1.585\\
4 & 0.236 & 1.449\\
5 & 0.258 & 1.374\\
\hline
\end{tabular}
\end{center}
\end{table}

Table \ref{tab:result_clustering} shows the clustering performance for 5 iterations of our proposed method. The clustering performance is always improving and this can provide more visually correlated merged food items for CNN model to learn.

\subsection{Qualitative results from merging food items}

Figure~\ref{fig:qual_item} presents a the outcome of merging food items within the "bacon" food type category. Upon observation, it becomes evident that all food items with visual similarity have been cohesively merged to form a newly merged food item. For instance, the food item identified by USDA food code 22311450, which originally has only three training images, is merged with the food item identified by USDA food code 24208500, which originally has one training image. As a result of this merging, we now have a training set of four images to train the model under a single merged food item. This is particularly beneficial in mitigating the class imbalance issue inherent in the data. Such a strategy highlights the efficiency of our approach in leveraging limited image data from specific classes for effective training. This, in turn, facilitates the CNN models in acquiring more discriminative image features.

\section{Conclusion}
In this paper, we exploited a VFN-nutrient dataset that labels each food image with a corresponding food item and its nutritional composition information. We consider food items as subcategories of food types, thereby forming a hierarchical structure within the dataset. Predicting food items enables us to retrieve the corresponding nutritional composition information, thus bringing us closer to the goal of image-based dietary assessment. To create visual relations among food items, we implemented an end-to-end food items merging method during training phase by updating CNN model for extracting image features iteratively through multi-stage hierarchical transfer learning, which can also address the class imbalance issue across food items.  

Despite these contributions, there are other potential strategies that could further improve food item classification accuracy. For future work, we aim to incorporate ideas from multi-modal learning to enhance our classification of food items, and in turn, refine the results of food image classification on food items.


\bibliographystyle{ACM-Reference-Format}
\balance
\bibliography{reference}


\end{document}